\documentclass[letterpaper]{article} 
\usepackage{aaai2027}  
\usepackage{times}  
\usepackage{helvet}  
\usepackage{courier}  
\usepackage[hyphens]{url}  
\usepackage{graphicx} 
\usepackage{natbib}  
\usepackage{caption} 
\usepackage{booktabs}
\usepackage{amsmath} 
\usepackage{algorithm}
\usepackage{algpseudocode}
\usepackage{placeins}

\usepackage{newfloat}
\usepackage{listings}
\DeclareCaptionStyle{ruled}{labelfont=normalfont,labelsep=colon,strut=off} 
\floatstyle{ruled}
\newfloat{listing}{tb}{lst}{}
\floatname{listing}{Listing}
\title{Personalized Execution Time Optimization for Billion-Scale Scheduled Jobs}

\author{ 
    Yang Liu\corresponding, Juan Wang, Idris Malik, Zhengxing Chen, Ian Fox,\\
    Imani Mufti, Jason Sukumaran, Baokun He, Xiling Sun, Feng Liang
}
\affiliations{
    Meta \\
    1 Hacker Way, Menlo Park, CA, USA, 94025 \\
    yliu9@meta.com, juanw@meta.com, idrismalik@meta.com, czxttkl@meta.com, ifox@meta.com, \\ imanimufti@meta.com, jsukumar@meta.com, baokunhe@meta.com, xlsun@meta.com, liangfeng@meta.com
}

\usepackage{xparse}
\makeatletter
\NewDocumentCommand{\LeftComment}{s m}{%
  \Statex \IfBooleanF{#1}{\hspace*{\ALG@thistlm}}\(\triangleright\) #2}
\makeatother

\usepackage{bibentry}

\begin{document}

\maketitle

\begin{abstract}
Scheduled batch jobs are widely used on asynchronous computing platforms to execute enterprise applications such as promotional notifications and candidate pre-computation for recommender systems. Delivering or updating information at the right time is important for user experience and execution impact, yet providing a versatile, personalized execution time optimization solution across diverse product scenarios while maintaining reasonable infrastructure costs remains challenging. In this paper, we present a deployed system that serves billions of users daily, combining learning-to-rank with a ``best time policy'' for execution time selection. We describe the four-stage evolution of our approach: from heuristic peak-hour rules, to pointwise ML-based activity pattern predictions, to a linear signal assembler with globally fixed weights, and finally to a contextual ensemble learner that produces per-user adaptive fusion weights via a neural policy network trained with listwise learning-to-rank objectives. We further report the discovery of cross-use-case cannibalization effects and introduce a coordination system to mitigate the problem. Our production experiments demonstrate measurable improvements in both execution efficiency and downstream product impact. We share deployment lessons including failure analyses and design decisions accumulated over four years of operating this system at scale. To our knowledge, this represents the first ML-based multi-tenant execution time optimization system deployed across different product domains at industrial scale. 
\end{abstract}

\section{Introduction}
A job scheduler or task queue system has been widely deployed in modern distributed computing architectures to ensure execution and delivery of tasks at billion-scale volume, including Airbnb Dynein \cite{schedulingairbnb}, Meta FOQS \cite{schedulingfacebook}, Google Cloud Task Queue \cite{schedulingoogle}, and AWS Job Queue \cite{schedulingaws}. The advantages of the scheduled tasks on the asynchronous computing architecture include the decoupling from user request, the reduced request latency, the better scalability and management with the rate-limiting, retry, and priority control \cite{schedulingoogle}. Scheduled jobs are particularly suitable for complex applications with tolerable delays, such as notifications, video encoding, and candidate pre-computation for recommender systems. Those scheduled applications can be personalized and executed at per user basis by providing the user information with its execution timestamps in the task queue via the job scheduler. In social networks, it is important to deliver or update information to users at the right time to maintain user experience and maximize execution impact. However, no mature personalized execution time optimization solutions for scheduled jobs have been reported in industrial applications. We address this gap with a deployed system that has been in production for over four years. 

Notifications are a primary channel for delivering information in social network products, and delivery time is an established factor in engagement \cite{pejovic2014interruptme} \cite{ho2018nurture}. Users receive over 65 notifications per day on average, and nearly half interrupt an ongoing task \cite{pielot2014situ}. Engagement depends on device context \cite{okoshi2017attention} and content relevance, but promotional notifications are scheduled in advance \cite{liu2020reinforcement}, so device-level and environment signals are unavailable at scheduling time. Beyond notifications, periodic candidate generation is widely used in billion-scale web applications \cite{agarwal2015personalizing} \cite{covington2016deep} \cite{linden2003amazon}, where infrastructure constraints make it challenging to refresh candidates at the right time---especially for the large population of non-daily active users. A reinforcement learning approach has been introduced to optimize the personalized number of executions \cite{liu2020reinforcement}; the remaining question, which we address, is how to choose the execution timestamps once that number is determined.

In this paper, we formulate the execution time optimization as a multi-level optimization problem: \begin{itemize} 
\item \textbf{Business objective}: Maximize user engagement while respecting notification volume budgets and infrastructure constraints. 
\item \textbf{Proxy metric}: Maximize the probability that the user is active (or responsive) at the selected execution time slots. 
\item \textbf{Learning objective}: Minimize listwise ranking loss over candidate time slots, ensuring the top-ranked slots align with actual user activity patterns. 
\item \textbf{System constraints}: Balance infrastructure load across time slots; avoid cannibalization across co-located use cases. 
\end{itemize} 

We encounter four major challenges. First, the best time selection problem demonstrates strong dependency among ranking objects---selecting one time slot affects the utility of others, similar to slate ranking \cite{bello2018seq2slate}. Second, a single-channel utility function does not fit all use cases, especially for channel-inactive users. Third, the number of execution times varies widely across product use cases (from once per week to 24 daily). Fourth, we observe cannibalization across use cases competing for peak time slots. 

We make the following contributions: 
\begin{itemize} 
\item Formulate execution time optimization as a learning-to-rank problem and propose a ``best time policy'' for scheduled job time selection. 
\item Propose and compare two ensemble approaches: a Linear Signal Assembler (LSA) with globally fixed weights, and a Contextual Signal Assembler (CSA), a neural ensemble trained with listwise learning-to-rank objectives. 
\item Report the discovery and mitigation of cross-use-case cannibalization effects. 
\item Present comprehensive production results from a system serving billions of users, including a 4-year deployment retrospective with lessons learned. 
\end{itemize}

To our best knowledge, our work has been the first ML-based algorithm to support the multi-tenant application to solve the execution time optimization problem for the scheduled jobs at industrial scale across different product domains.

\section{Related Work}
\label{related_work}

Prior work on notification timing separates from ours along two axes. On the \textit{phase} axis, email marketing services distribute mail at optimal times \cite{kalantari2021trackers} and Pinterest's budget pacer schedules at user peak hours \cite{zhao2018notification}, but both expose only a few peak slots per day and cannot satisfy diverse scheduling needs; LinkedIn's notification
optimization \cite{gao2018near} and email send/drop decisions \cite{gupta2016email, gupta2017optimizing} filter at the delivery phase rather than optimizing the schedule itself. On the \textit{signal} axis, timing work using heuristic rules \cite{fischer2011investigating}, supervised learning
\cite{pejovic2014interruptme}, reinforcement learning \cite{ho2018nurture}, device-side breakpoint detection \cite{okoshi2019realworld}, and offline reinforcement learning \cite{yuan2022offline, sutton2019reinforcement} all
operate on the device with real-time context, which is unavailable when promotional jobs are scheduled a day in advance. Complementary efforts optimize the \textit{volume} dimension under constraints
\cite{zhang2023volume} or model the future value of a send to decide send/no-send \cite{obrien2022should}, and reinforcement learning has been proposed for omnichannel channel selection
\cite{oliveira2026hyperpersonalization}; none reports a deployed system for scheduling-phase execution time optimization at scale.

Learning-to-rank has been widely applied in information retrieval and recommender systems~\cite{chapelle2011future}, but most applications implicitly assume independence among ranking objects. The best time selection problem exhibits strong inter-slot dependencies: user activity patterns are temporally correlated (a user active at 9pm is likely also active at neighboring hours), and selecting one time slot causally affects subsequent slots---delivering a notification at 9pm diminishes the user's responsiveness at 10pm due to attention saturation. Prior work has addressed inter-object dependencies through slate ranking~\cite{bello2018seq2slate, ie2019slateq}, diversity-aware ranking~\cite{yan2021diversification}, conditional random fields~\cite{qin2008global}, and sequential MDP formulations~\cite{zhao2018deep}. However, modeling the conditional probability structure among time slots at scale is infeasible when the number of execution times varies from once per week to 24 daily. We instead propose a greedy policy-based approach (Best Time Policy section) that heuristically captures these dependencies without explicit joint probability modeling.

Feature-weighted linear stacking combines model predictions with static
weights \cite{sill2009feature}; mixture-of-experts \cite{shazeer2017moe} and
contextual model selection \cite{ma2018mmoe} learn combination weights
conditioned on input features. Our contextual assembler (Contextual Signal
Assembler Section) extends this line by learning per-user fusion weights over
temporal activity sources with a listwise objective, replacing linear
stacking's globally fixed weights with context-aware source weighting.

\section{Methodology}
\label{methodology}
\subsection{Problem Formulation}
Given a job to be executed within a time range, the task is to identify the $N$ time slots that best optimize product-specific downstream metrics, at a marginal computational cost low enough to serve hundreds of use cases from one service. That constraint determines where the learning sits: modeling capacity is mainly spent on a shared understanding of when each user is active, with a lightweight per-use-case fusion layer on top, while the per-use-case decision layer remains a configurable rule applied at serving time. The optimizer takes a scheduling request \textit{r} containing the product use case \textit{p}, user \textit{u}, execution time range $[t_{start}, t_{end}]$, number of time slots $N_{u,p}$, slot length \textit{I}, prediction metric \textit{m}, and best time policy $\pi$. Note that the execution frequency $N_{u,p}$ is determined by the upstream scheduling policy \cite{liu2020reinforcement} and is not optimized here; our task is to select \textit{which} $N$ time slots maximize the per-execution engagement. We define \textbf{execution efficiency} as the product-specific engagement metric (\textit{e.g.}, clicks, sessions) aggregated over all $N_{u,p}$ executions within the day, normalized by $N_{u,p}$. This metric measures how effectively each scheduling event converts into user engagement---higher efficiency indicates better alignment between the selected time slots and the user's activity patterns. We define \textbf{execution impact} as the relative change in a product-specific, user-level daily engagement metric between the optimized group and the baseline.

\subsection{Time Slot Ranking}
Given slot length \textit{I}, the execution time range is partitioned into contiguous candidate time slots $\mathcal{T} = \{t_1, t_2, \ldots, t_k\} \; (k \geq N)$. All time slots are defined in a single server-side timezone rather than each user's local timezone, for two practical reasons: reliable timezone attribution is not available for all users and daylight saving transition dates differ across countries; and a single reference frame keeps logging, training, and serving consistent, avoiding training--serving skew. This does not limit personalization---a user's peak activity simply appears at a different server-side hour depending on their location, and the utility functions learn this per-user offset directly from the data. Section Deployment Lessons describes a failure this design choice exposed. We create a user temporal activity map $V_{u,m}$ that maps each time slot $t$ to its personalized predicted value, generated by a learned utility function $f_m(u,t)$.

The optimizer returns a slot, not a wall-clock timestamp. The job scheduler draws the actual execution time uniformly within the returned slot,
using a per-use-case jitter configuration, so a slot of 8am places the job in the delivery queue somewhere in 8:00--8:59am. This is what keeps billions of daily executions from collecting at the top of the hour, and it sets the granularity at which timing is optimized: the utility functions and the best time policy both operate on slots, and the paper makes no claim about sub-slot precision.

\subsection{Utility Functions}
We provide utility functions $f_m$ in two categories:

\begin{itemize}
    \item pClick models: Classification models predicting notification click probability at time slot \textit{t}. 
    \item Activity models: Regression models predicting user activity metric $m$ at time slot \textit{t}.
\end{itemize}

These scores are normalized to $[0.0, 1.0]$ by taking 
\begin{equation} 
\label{eq:norm} \widetilde{m_{u,t}} = \frac{m_{u, t} - \min(m)}{\max(m) - \min(m)} \end{equation} where $\max(m)$ and $\min(m)$ are the maximal and minimal prediction value of metric $m$ across all users and time slots, respectively. Normalization serves two purposes. First, the sources differ in units---the CTR model emits a probability, while the remaining sources emit time-spent in seconds---so Eq.~\ref{eq:norm} places them on a common scale where the fusion weights in Eqs.~\ref{eq:lsa} and \ref{eq:csa} are comparable. Second, as a per-signal monotone affine map it preserves each source's within-user temporal ordering, rescaling sources relative to one another without changing the ranking any single source induces on its own; the learned fusion weights absorb any residual scale differences across sources.

\subsection{Linear Signal Assembler (LSA)}
\label{sec:lsa}
In a multi-tenant optimization service, it is infeasible to provide a dedicated utility function for every product use case given limited infrastructure resources. Moreover, the model quality of product-specific utility functions degrades over time due to the sparsity of positive labels. We therefore provide a variety of generic utility functions covering user activity on different delivery channels, and combine them via a product-specific signal assembler to generate a unified temporal activity score $M_{u,t,p}$ for each use case $p$:

\begin{equation} 
\label{eq:lsa}
M_{u,t,p}^{(LSA)} = \sum_{m \in \mathcal{M}} \omega_{m,p} \times \mathcal{A}_{u,m} \times \widetilde{m_{u,t}} 
\end{equation} 

where $\omega_{m,p}$ are the product-specific metric weights, $\mathcal{A}_{u,m}$ is the user's activity indicator on the channel associated with signal $m$, and $\widetilde{m_{u,t}}$ is the normalized prediction score. 
The weights $\omega_{m,p}$ are learned offline by minimizing a rank-squared loss via feature-weighted linear stacking  \cite{sill2009feature}, and the learned weights are then applied globally to all users within the same use case. 

\subsection{Contextual Signal Assembler (CSA)} 
\label{sec:csa} 
Operating LSA at scale for over two years revealed three limitations: (1) static global weights cannot capture user-context heterogeneity; (2) manual weight tuning does not scale to hundreds of use cases; and (3) the rank-squared training surrogate does not directly optimize listwise ranking quality. Drawing on ideas from adaptive mixture-of-experts~\cite{jacobs1991adaptive, shazeer2017moe} and contextual model selection~\cite{ma2018mmoe}, we introduce a contextual ensemble learner that addresses all three issues by producing \textit{per-user} adaptive fusion weights conditioned on user context.

\subsubsection{Architecture.} 
The model takes three inputs: (1) a user context feature vector $\mathbf{x}_u$ encoding demographics, activity level, and platform; (2) per-source temporal activity predictions $\{\widetilde{m_{u,t}^{(s)}}\}_{s=1}^{S}$ from $S$ signal sources; (3) a source availability mask $\mathbf{a}_u \in \{0,1\}^S$ indicating which signals are available for user $u$ (not all users have activity on all channels). A multi-layer perceptron (MLP) policy network $g_\theta$ maps the user context and availability mask to per-source fusion weights: 
\begin{equation} 
\label{eq:softmax}
\mathbf{w}_u = \text{Masked Softmax}(g_\theta(\mathbf{x}_u, \mathbf{a}_u), \; \mathbf{a}_u)
\end{equation} 
where the masked softmax assigns zero weight to unavailable sources and normalizes the remaining weights to sum to one. This allows a single model to gracefully handle all source-availability patterns without requiring separate models per combination. The fused temporal activity score is then computed as: 
\begin{equation} 
\label{eq:csa}
M_{u,t,p}^{(CSA)} = \sum_{s=1}^{S} w_u^{(s)} \times \widetilde{m_{u,t}^{(s)}} 
\end{equation} 
Unlike LSA's global weights, CSA's fusion weights vary per user: a user with rich push notification history may weight that signal heavily, while a user active primarily on in-app channels receives a different combination---all learned automatically from data. 

\subsubsection{Training Objectives.} 
A key insight is that the downstream task—selecting the best \textit{N} time slots—depends on the relative ranking of time slots, not their absolute predicted values. The pointwise utility functions are trained with per-hour binary cross-entropy or squared error loss, which optimizes prediction accuracy for each time slot independently. However, minimizing per-hour prediction error does not guarantee correct ranking of peak hours: a model can achieve low MSE while reversing the order of similarly-valued adjacent time slots. The CSA addresses this objective mismatch by directly optimizing listwise ranking quality over the fused scores, ensuring the top-ranked slots align with the user’s true peak activity hours. We implement three listwise learning-to-rank algorithms and compare their effectiveness: 

\begin{enumerate} 
\item \textbf{Surrogate (ListNet-style)} \cite{cao2007listnet}. We use cross-entropy loss between the fused score distribution (converted to a probability via softmax over time slots) and the observed future user activity distribution. This serves as a differentiable listwise proxy for ranking quality, analogous to ListNet's top-one probability framework. 
\item \textbf{REINFORCE-based direct metric optimization.} We directly maximize NDCG@K---the true non-differentiable ranking metric---using the REINFORCE algorithm \cite{williams1992simple} with a Dirichlet stochastic policy over fusion weights. This bypasses the surrogate's proxy-metric gap through policy gradient estimation. 
\item \textbf{Evolution Strategies (ES)} \cite{salimans2017evolution}. A gradient-free alternative that optimizes NDCG@K via antithetic perturbation sampling. ES requires no backpropagation through the ranking metric and is robust to non-smooth reward landscapes. 
\end{enumerate} 

\subsection{Best Time Policy}
\begin{figure}[t] \centering \includegraphics[width=\linewidth]{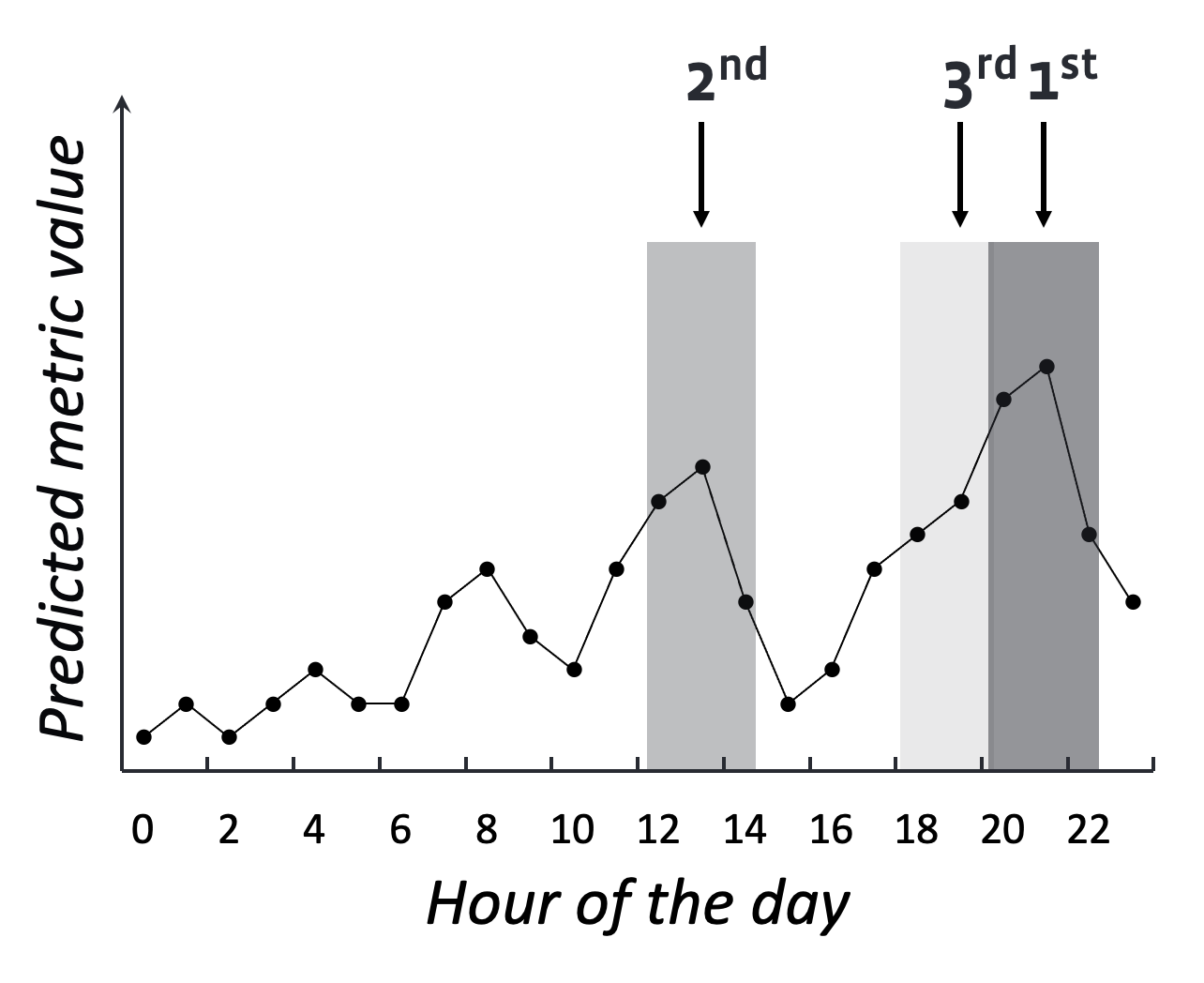} \caption{An example of applying avoid 1-hour nearby best time policy to return three execution timestamps. After the current peak hour is selected, the peak hour and its nearby hours are removed from the temporal activity map $V_{u,m}$, from where the next peak hour is selected. With $N = 3$, the policy returns 9pm, 1pm, and 7pm in server-side time.} \label{fig:besthourpolicy} \end{figure}

We observe that selecting the top-N time slots ignores inter-slot dependencies and generally does not result in optimal performance (Table ~\ref{besthourpolicy-table}). We propose an explainable yet robust technique called ``Best Time Policy'' on top of the utility functions. These policies return the best time slots based on the provided temporal activity map and the number of best time slots. Note that for $N = 1$ the avoid-nearby policy reduces to top-1 selection, since the avoidance window only constrains subsequent selections. Figure~\ref{fig:besthourpolicy} illustrates the avoid nearby policy and its implementation is documented in Algorithm~\ref{algorithm:avoidnearby}.

Clients configure $w$ and $n$ per request (Algorithm~\ref{algorithm:avoidnearby}), making inter-slot spacing a per-use-case setting rather than a global constant. Our optimizer can be further extended to support more use case-specific best time policy based on the product context. 

\begin{algorithm}[t]
	\caption{Avoid $w$-slot nearby best time policy} 
	\begin{algorithmic}[1]
    \Function{$\mathcal{\pi}$}{$p, V_{u,m}, n, w$} 
    \LeftComment{Return $n$ best time slots with $w$ avoidance window}
        \State $best\_time\_list \gets \emptyset$
        \If {use case $p$ is low priority}
		    \State \textbf{remove} top peak time slot from $V_{u,m}$
		\EndIf
        \While{$n > 0$}
            \State $n\gets n - 1$
            \State \textbf{identify} top time slot $best\_time$ from $V_{u,m}$
            \State $best\_time\_list$.append($best\_time$)
            \For{\texttt{$t : m_{u,t}$ in $V_{u,m}$}} 
            \LeftComment {\%Remove peak and nearby slots\%}
                \If {$t \geq best\_time - w$ and $t \leq best\_time + w $}
		            \State \textbf{remove} $t : m_{u,t}$ from $V_{u,m}$
                \EndIf
            \EndFor
        \EndWhile\label{euclidendwhile}
        \State \textbf{return} $best\_time\_list$
    \EndFunction
    \end{algorithmic}
    \label{algorithm:avoidnearby}
\end{algorithm}

\subsection{Best Time Coordination}
Cannibalization is a general platform-level issue and we ask whether different use cases in our optimization service compete for the user's peak time slot and impair the global delivery impact. We classified the use cases into two tiers based on their business context. In the low priority group, we remove the top peak time slot from the candidate time slots $\mathcal{T}$ during the best time policy implementation.

\section{Applications and Experiments}
\label{experiments}
\subsection{Architecture}
The system consists of five components: (1) a data pipeline collecting ML training data; (2) weekly recurring model training; (3) offline model inference pipelines that generate per-source user temporal activity predictions; (4) an offline ensemble learning pipeline that fuses per-source predictions into a unified temporal activity score, making the system compositional and transferable across use cases; and (5) an online execution time optimization service that applies best time policy and coordination to return execution timestamps to the job scheduler. Both per-source and fused scores are stored in key-value storage services on a day-of-week basis, so a request resolves to a lookup rather than an inference. Over a representative week the service averages under $50$\,ms per request, tolerable for a day-ahead scheduling path, and meets its SLA on $99.999\%$ of calls.

The signal assembly step—combining multiple prediction sources into a unified temporal activity score—is performed offline for both LSA and CSA. The system provides five categories of user temporal activity signals, comprising six input sources ($S = 6$; the aggregated activity pattern contributes separate 7-day and 28-day variants):
\begin{enumerate} 
\item Time-spent prediction: per-hour time-spent prediction from a pointwise regression model.
\item CTR prediction: per-hour notification click-through rate prediction conditioned on delivery, from a pointwise classification model.
\item Aggregated activity pattern: per-hour time-spent aggregated from the user’s recent 7-day and 28-day history.
\item Graph-based pattern: per-hour time-spent pattern derived from the user’s social network neighbors.
\item Segment-based pattern: per-hour time-spent pattern from the user’s behavior-based segment \cite{liu2026bus}.
\end{enumerate} 

\begin{figure*}
\includegraphics[width=\textwidth]{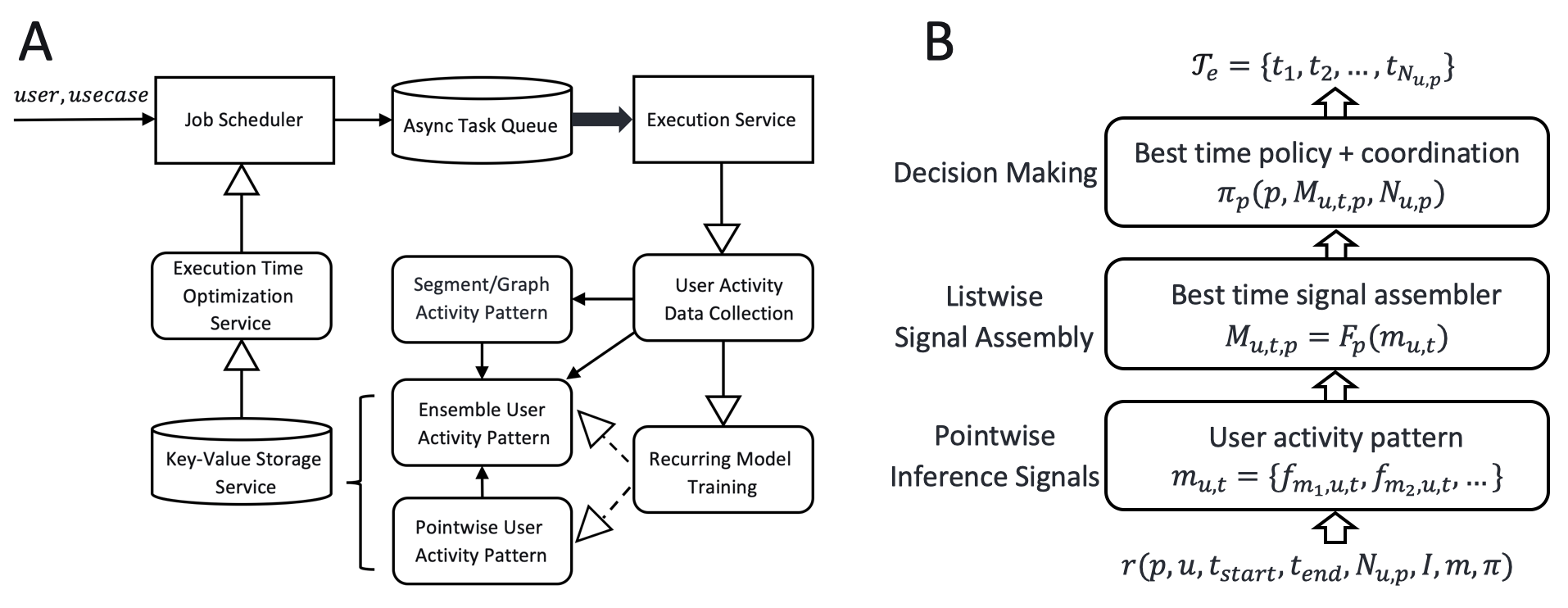}
\caption{System Overview. (A) End-to-end flow across the five components described in the Architecture section. (B) The \textit{Pointwise Inference Signals} layer harbors various user activity pattern signals and each signal represents the ranking score of each time slot from a single data source. The \textit{Listwise Signal Assembly} layer aggregates various user activity patterns to generate a use case-specific representation. The \textit{Decision Making} layer applies a use case-specific best time policy and platform-level coordination to finish the time slot ranking.}
  \label{fig:overview}
\end{figure*}

In LSA, these sources are combined via a static linear weighted sum with globally fixed weights. In CSA, a contextual MLP produces per-user adaptive fusion weights conditioned on user features and source availability. In both cases, the fused temporal activity score is written to the key-value store and consumed by the online optimization service, which applies the best time policy and coordination to return execution timestamps to the job scheduler (Figure~\ref{fig:overview}).

A natural alternative is to incorporate all signal sources as input features to a single pointwise or listwise ranking model. We chose the ensemble architecture for four practical reasons: (1) multi-tenancy—hundreds of use cases with different optimization objectives can customize their signal combination without retraining shared models; (2) heterogeneous update schedules—different signals have different freshness requirements (daily counters vs. weekly model predictions vs. monthly graph features), and ensemble fusion decouples their maintenance; (3) variable source availability—not all users have all signals available, and the masked softmax handles missing sources without imputation; and (4) production stability---when a source degrades or its pipeline is delayed, the ensemble gracefully down-weights the affected signal rather than failing entirely, making it more fault-tolerant than a monolithic end-to-end model. This offline design reduces online serving latency and decouples the ensemble model's update cycle from the online service's release process. 

\subsection{Pointwise Model Training} 
The training data were collected from randomly sampled 1\% production traffic and daily-shuffled small exploration traffic ($\le$ 0.1\%) in the best time decision making. The training samples were joined with server-side tracking data to generate per-hour activity labels for user $u$ on time slot $t$. We trained two categories of pointwise models using Light Gradient Boosting Machine (LightGBM) \cite{ke2017lightgbm}: \begin{itemize} 
\item \textbf{pClick model (classification).} Each training row corresponds to an actual send attempt; the label is 1 if the user clicked within one hour of delivery and 0 otherwise. Because the production policy concentrates deliveries at predicted peak hours, we combine two streams: a randomized-timestamp exploration stream that provides unbiased coverage of all hours, and a sampled production stream that excludes exploration users. We downsample each hour bucket to the size of the smallest bucket, yielding uniform coverage across the 24 candidate slots. Negatives are additionally downsampled and the per-row sampling rate is retained for calibration. The model is trained with logistic loss to predict $P(\text{click} \mid u, t)$. Note that the activity model below is not subject to this exposure bias, as time-spent is observed unconditionally on delivery.
\item \textbf{Activity model (regression).} The label is the user's time-spent (in seconds) at time slot $t$. Due to the long-tailed distribution of time-spent, we applied stratified sampling to balance the training data and a natural log transformation to normalize the regression labels. A reverse (exponential) transformation is applied to the prediction scores in the inference pipeline. The model is trained with squared error loss to predict the expected time-spent $\hat{m}_{u,t}$. \end{itemize} Both models are trained on the internal ML platform \cite{wu2019machine} with weekly recurring updates to capture evolving user behavior patterns.

\subsection{Ensemble Learning}
The CSA training pipeline constructs per-user training examples by joining multiple signal sources with ground-truth engagement labels. Specifically, the training data are generated through the following steps: 
\textbf{Source assembly.} For each user $u$, we collect the hourly temporal activity predictions from all $S$ signal sources in Architecture Section and aggregate them into a per-user source label map $\{\text{source}_s \rightarrow \widetilde{m_{u,t}^{(s)}}\}_{s=1}^{S}$. Training data are sampled at 1\% of production traffic to maintain computational tractability while preserving population-level diversity. \textbf{User features.} We extract 12 user-level features from the user attribute table, including activity, social graph features, and demographic features. Categorical features are encoded via consistent hashing to produce dense numerical representations without requiring a lookup table. \textbf{Ground-truth labels.} The target labels $y_{u,p,t}$ are the user's hourly engagement aggregated over a three-day forward-looking window following the prediction date. The choice of label is use-case-specific and represents the hourly engagement metric that each use case aims to optimize. This flexibility allows each use case to define its own optimization objective while reusing the same shared signal building blocks. \textbf{Training procedure.} The assembled training data---user features $\mathbf{x}_u$, source predictions $\{\widetilde{m_{u,t}^{(s)}}\}_{s=1}^{S}$, and target labels $\{y_{u,p,t}\}_{t=1}^{24}$---are fed to the MLP policy network. Training-set size depends on the label: 10M users for the notification CTR label and 25M users for the time-spent label. The model is trained using the Listwise Cross-Entropy Loss (ListNet surrogate) with the following configuration: 1 node $\times$ 4 GPUs (Grand Teton T20), batch size of 512, and learning rate of 1e-3 with cosine annealing. The model is retrained weekly on refreshed labels to incorporate recent behavioral shifts. Concretely, the objective is the cross-entropy between the predicted distribution over time slots and the user's observed engagement distribution: 

\begin{equation} \mathcal{L}_{\mathrm{CSA}} = -\frac{1}{|\mathcal{U}|}\sum_{u \in \mathcal{U}} \sum_{t=1}^{|\mathcal{T}|} P_{\mathrm{true}}(t \mid u) \cdot \log P_{\mathrm{pred}}(t \mid u) \end{equation} 

where $P_{\mathrm{true}}(t \mid u) = y_{u,t} / \sum_{t'} y_{u,t'}$ is the normalized ground-truth distribution and $P_{\mathrm{pred}}(t \mid u) = \exp(M^{(\mathrm{CSA})}_{u,t}) / \sum_{t'} \exp(M^{(\mathrm{CSA})}_{u,t'})$ is the softmax over fused scores. Training-resource comparisons across objectives (Table~\ref{tab:objective-comparison}) use a single data snapshot and package version.

\textbf{Inference.} The model is then deployed via an offline inference pipeline that processes the full population of billions of users at a daily computational cost of 125 Batch CPU Units (BCU), within the 100--150 BCU range of the LSA's aggregation pipeline---the resource use is dominated by reading per-source predictions and writing fused scores, which both approaches share.

\subsection{Offline Evaluation} 
We evaluate the ranking quality of the activity pattern signals using NDCG \cite{jarvelin2002cumulated}. For each user, the 24 hours of the day are ranked by the signal's predicted score and compared against the user's actual hourly engagement pattern. We evaluate against two ground-truth targets: (i) next-day in-app time-spent, which reflects general user activity, and (ii) next-7-day notification conversion count, which reflects channel-specific responsiveness. 
\subsubsection{Overall Ranking Quality.} Table~\ref{tab:offline-ndcg} reports the offline NDCG for the key signals in our system. Three findings emerge. First, CSA trained with time-spent label achieves the best ranking quality on the time-spent target (NDCG@1 = 0.261), outperforming both the pointwise timespent model (0.258) and the LSA ensemble (0.254). Second, CSA trained with notification CTR label achieves the best ranking quality on the notification conversion target (NDCG@1 = 0.167). Third, each pointwise signal ranks best on the target it was trained for. On the notification conversion target, the pointwise CTR model (0.161) outperforms the pointwise time-spent model (0.155); on the time-spent target the ordering reverses decisively (0.258 vs. 0.168). This label--target alignment validates the use-case-specific label design in CSA.

\begin{table}[t] 
\centering 
\setlength{\tabcolsep}{1mm} 
\begin{tabular}{l|ccc|ccc} \toprule & \multicolumn{3}{c|}{Time-spent target} & \multicolumn{3}{c}{Notif. conv. target} \\ Signal & @1 & @5 & @24 & @1 & @5 & @24 \\ \hline CSA ts & \textbf{.261} & \textbf{.366} & \textbf{.580} & .160 & .260 & .493 \\ CSA ctr & .253 & .357 & .575 & \textbf{.167} & \textbf{.271} & \textbf{.501} \\ LSA ts & .254 & .358 & .575 & .158 & .259 & .493 \\ Pointwise ts & .258 & .363 & .578 & .155 & .253 & .489 \\  Pointwise ctr & .168 & .270 & .522 & .161 & .270 & .500 \\ \hline Peak hours & .166 & .260 & .508 & .138 & .226 & .467 \\ Pop.\ average & .150 & .239 & .493 & .132 & .220 & .462 \\ \bottomrule \end{tabular} \caption{Offline NDCG against two ground-truth targets, evaluated on a random sample of users with non-empty ground-truth activity in the evaluation window. Bold indicates the best score among non-baseline signals. CSA with time-spent label leads on the time-spent target; CSA with CTR label leads on the notification conversion target.} \label{tab:offline-ndcg} \end{table} 

\subsubsection{User Activity Cohort Analysis.} 
\begin{figure}[t]
\centering
\includegraphics[width=0.78\linewidth]{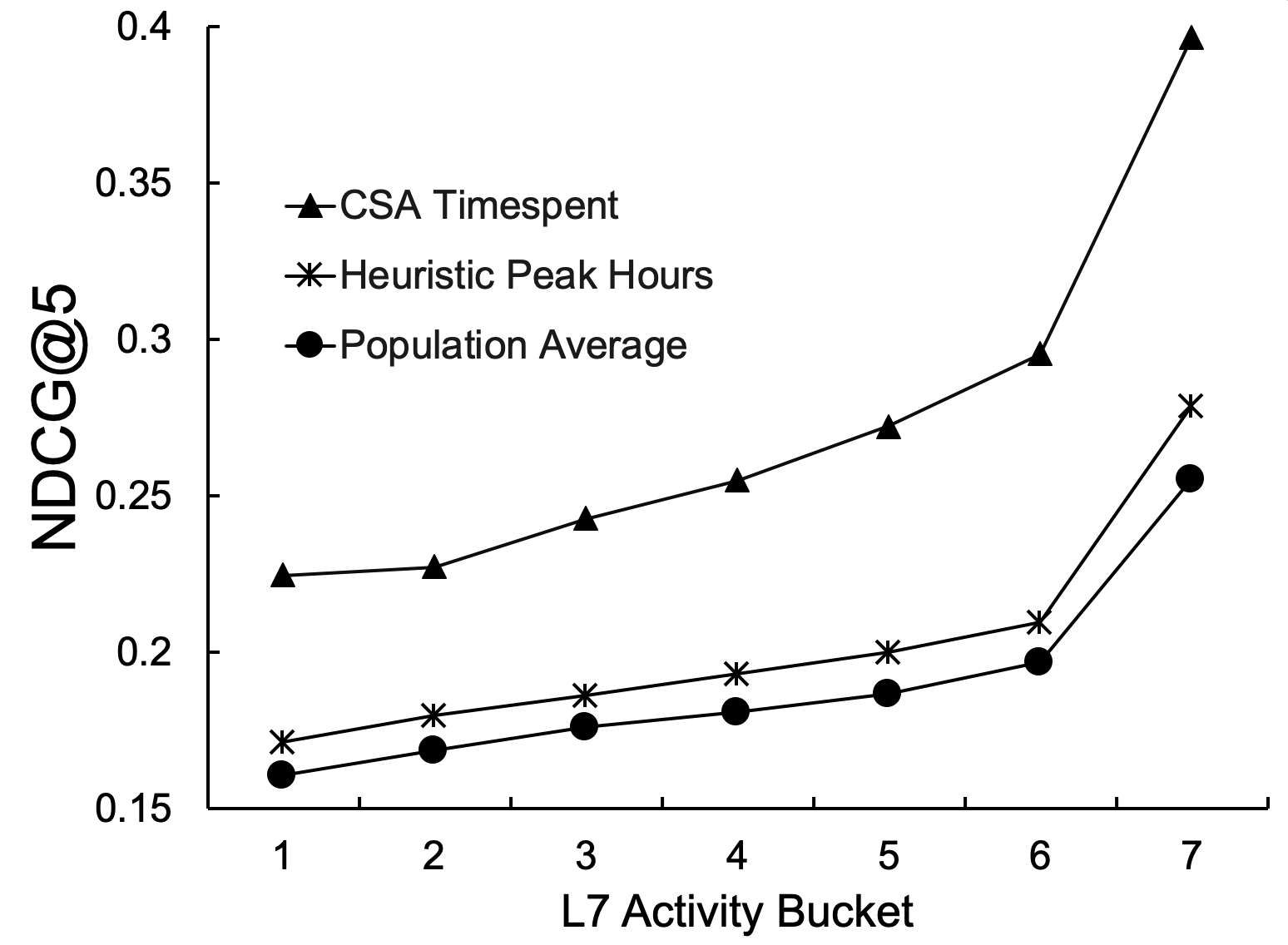}
\caption{User activity cohort analysis (L7, the number of days active in the last 7 days) of the ranking quality (NDCG@5 vs.\ next-day time-spent). Ranking quality increases monotonically with user activity for all signals. CSA time-spent consistently leads, with the largest margin on the most active cohort.}
\label{fig:offline}
\end{figure}

We further analyze ranking quality by user activity level (Figure~\ref{fig:offline}). NDCG@5 is positively correlated with user daily activity across all signals: for the most active cohort (L7 = 7), CSA time-spent achieves NDCG@5 = 0.396, while for the least active cohort (L7 = 1), it drops to 0.225. Active users exhibit more regular, easier-to-predict temporal patterns. Importantly, for low-activity users, the gap between individual signals narrows and all signals approach the population average baseline, indicating near-random ranking quality from any single source. This motivates the CSA's multi-source fusion: by combining heterogeneous signals---including graph-based and segment-based patterns that provide coverage for inactive users---the CSA compensates where channel-specific models alone cannot provide reliable temporal predictions. 

\subsubsection{Training Objective Comparison.} 
\begin{table}[t] 
\centering 
\setlength{\tabcolsep}{1mm}
\small \setlength{\tabcolsep}{4pt} \begin{tabular}{l|ccccc} 
\toprule 
Objective & Best    & Best & Epochs & GPU-h & GPU-h \\
& NDCG@5  & ep.  & run    & best  & full  \\
\midrule
Surrogate & \textbf{0.2407} & 5   & 31  & \textbf{2.1} & \textbf{13.4} \\
REINFORCE & 0.2367          & 31  & 41  & 15.5 & 21.0 \\
ES        & 0.2386          & 200 & 200 & 90.2 & 90.7 \\\end{tabular} \caption{Comparison of CSA training objectives; architecture, data, and
hardware are identical across runs (see text). \textit{Best ep.} is the epoch
attaining the peak NDCG@5; runs continue past it until early stopping, so \textit{Epochs run} reflects the full run rather than \textit{Best ep.}
Surrogate and REINFORCE stopped early; ES reached its 200-epoch cap without
early stopping, so its GPU-hour cost is a lower bound. GPU-h is wall-clock
$\times$ 4 GPUs and includes $\sim$0.3--0.5 GPU-h of job startup. NDCG@5 is
computed on a held-out portion of the training batch against the three-day
training label, and is therefore not comparable to Table~\ref{tab:offline-ndcg}.} \label{tab:objective-comparison} \end{table}

Table~\ref{tab:objective-comparison} compares the three listwise training objectives under identical architecture and hardware (MLP with two 128-unit hidden layers, 6 sources, 12 user features, 1 node $\times$ 4 Grand Teton GPUs). The surrogate (ListNet-style) objective achieves the highest validation timespent pattern NDCG@5 (0.2407) at epoch 5, reaching 0.2401 by epoch 2 ($\sim$0.8 GPU-hours)---already above the best score either alternative attains at any point in training. Training the full run to early stopping costs 13.4 GPU-hours; the best checkpoint itself is reached in 2.1. REINFORCE peaks lower at 0.2367 (epoch 31, 21.0 GPU-hours for the full run). Evolution Strategies consumes its entire 200-epoch budget (90.7 GPU-hours) and is still improving at the cap, yet peaks lower at 0.2386. We adopt the surrogate objective for production deployment on computational cost grounds: under a fixed 200-epoch budget it attains the highest NDCG@5 of the three at roughly one-seventh the full-run GPU cost of ES. Because ES had not converged at the cap, its ceiling under a larger budget is unknown---but reaching it would cost at least 90.7 GPU-hours, a trade-off we consider unfavorable for a weekly-retrained production pipeline.

\subsection{Online Experiments} 

More effective timing could increase perceived intrusiveness if left unchecked; across all experiments above, production monitoring includes
opt-out and notification-disable rates as guardrail metrics alongside engagement, and a regression in either would halt rollout regardless of efficiency gains. All online experiments use user-level randomization and run long enough to cover whole weeks, so that day-of-week seasonality is balanced across arms. Reported intervals are 95\% confidence intervals produced by the internal experimentation platform, which winsorizes at the 99.95th percentile and then applies regression adjustment on pre-experiment data (equivalent to CUPED~\cite{deng2013cuped}).

\subsubsection{ML-based vs. Heuristic Baseline} We compared the ML-based optimizer against a heuristic rule-based baseline in A/B tests across seven production use cases. In the control group, the heuristic selects the top $N$ peak hours from the aggregated activity pattern based on the user's historical in-app activity. In the treatment group, the ML-based optimizer selects time slots using the learned utility functions and best time policy. Both arms schedule the same number of executions per user, so the comparison
isolates slot selection from volume: the heuristic fixes $N$ and the optimizer selects $N$ slots from the learned utility. Table~\ref{result-table} reports per-execution efficiency gains across four notification channels and scheduling frequencies from $N$ = 1 to 6. The ML-based approach improves efficiency for push notifications (+8.2\%, +8.1\%), red alert notifications (+4.6\%), SMS notifications (+3.1\%), and candidate pre-computation (+2.2\%, +2.1\%). We did not observe a measurable impact on email notifications ($case_7$). The comparison is against a heuristic peak-hour rule, so the null result says that we could not detect a meaningful gain from the learned utility on this channel, not that email delivery time is irrelevant---likely because users batch-process email at self-chosen times rather than responding to individual arrival times.
 
\begin{table}[t]
\centering
\small
\begin{tabular}{l|ccc}
\toprule
Scheduled job type & Efficiency (\%) & $N$ & Policy\\
\hline
$case_1$ push notif.       & +8.2$\pm$0.7 & 4 & avoid 1-h\\
$case_2$ push notif.       & +8.1$\pm$0.5 & 4 & avoid 1-h\\
$case_3$ red alert notif.  & +4.6$\pm$0.2 & 3 & avoid 2-h\\
$case_4$ precomputation    & +2.2$\pm$0.2 & 6 & avoid 1-h\\
$case_5$ precomputation    & +2.1$\pm$0.1 & 6 & avoid 1-h\\
$case_6$ SMS notif.        & +3.1$\pm$0.5 & 1 & top-N\\
$case_7$ email notif.      & n.s. & 3 & avoid 1-h\\
\bottomrule
\end{tabular}
\caption{A/B test results: per-execution efficiency gain of the ML-based approach over a heuristic rule-based baseline. $N$ denotes the number of scheduled time slots per day, and Policy denotes the best time policy applied in each use case. Values after $\pm$ are 95\% CI half-widths. All gains have $p < 0.05$; $case_7$ did not reach $p < 0.05$ (n.s.) and its interval is not available.}
\label{result-table}
\end{table}

\subsubsection{Best Time Policy Experiment}
We further made a case study and compared various best time policies in terms of execution efficiency improvement and product impact boosting. We observed that the avoid nearby policies have better performance than the top-N policy (Table ~\ref{besthourpolicy-table}). Based on these results, the avoid 1-hour and avoid 2-hour nearby policies are deployed as the production defaults for use cases scheduling more than one execution per day ($N > 1$).

\begin{table}[t] \centering \begin{tabular}{l|cc} \toprule Best time policy & Efficiency (\%) & Impact (\%) \\ \hline Top-N & control & control \\ Avoid 1-hour nearby & +2.10$\pm$0.74 & +0.110$\pm$0.072 \\ Avoid 2-hour nearby & +2.20$\pm$0.76 & +0.150$\pm$0.083 \\ Avoid 3-hour nearby & +1.65$\pm$0.59 & +0.125$\pm$0.079 \\ \bottomrule \end{tabular} \caption{Best time policies on a red alert notification use case with $N = 3$. Impact is the relative change in daily active users of the product surface driven by this use case. Values after $\pm$ are 95\% CI half-widths; all results have $p < 0.05$.} \label{besthourpolicy-table} \end{table}

\subsubsection{Contextual Signal Assembler Experiment.} 
We conducted an A/B test on SMS notification scheduling to evaluate the Contextual Signal Assembler (CSA). The existing production system uses a single pointwise notification CTR prediction to select the execution time. We compared three treatment arms against this baseline: the Linear Signal Assembler (LSA), CSA with in-app time-spent label, and CSA with notification CTR label. Table~\ref{tab:csa} reports the results. 

\begin{table}[t] \centering \begin{tabular}{l|cc} \toprule Method & Efficiency (\%) & Impact (\%)\\ \hline Pointwise CTR & control & control\\ LSA & +2.01$\pm$0.98 & +0.049$\pm$0.040 \\ CSA (TS label) & \textbf{+3.36}$\pm$0.66 & +0.068$\pm$0.061 \\ CSA (CTR label) & +3.22$\pm$1.02 & +0.057$\pm$0.041 \\ \bottomrule \end{tabular} \caption{A/B test on SMS notification scheduling, which schedules one execution per day ($N = 1$); all arms therefore select the top-ranked slot, isolating the effect of signal assembly. Baseline: pointwise CTR model. Impact is the relative change in app-level daily active users. Values after $\pm$ are 95\% CI half-widths. Every arm improves on the baseline at $p < 0.05$. In the direct contrast between CSA (TS label) and LSA, the efficiency gain reaches $p < 0.05$ while the impact difference does not.} \label{tab:csa} \end{table} 

Three findings emerge. First, both CSA and LSA outperform the single pointwise model, validating multi-source ensembling---LSA achieves +0.049\% DAU improvement, and both CSA variants improve further. Second, CSA with the time-spent label gives the largest point estimates on both metrics ($+3.36\%$ efficiency, $+0.068\%$ app-level daily active users). Against LSA directly, its efficiency gain reaches $p < 0.05$ while the impact difference is directionally consistent but does not: the $0.019$ percentage-point gap is well inside the interval this experiment resolves at app level, so the per-execution metric is where an assembly change is detectable at this traffic allocation. Third, the best label for SMS scheduling is the in-app time-spent pattern rather than the notification channel CTR---the timing signal that best predicts \textit{when the user will use the app} outperforms the signal that predicts \textit{when the user will click a notification}. This non-obvious finding underscores the importance of use-case-specific label selection.

\subsubsection{What the fusion model learned.} 
We inspected the per-user weights produced by the CSA (timespent label) on a 100K-user sample. The dominant source varies with user activity level: for the least active users (L28 1--3), the segment-based pattern receives 50\% of the weight, compensating for sparse individual history; for the most active users (L28 22--28), the 28-day time-spent pattern dominates at 36\%. Notably, controlling for activity level, users with high push notification engagement receive the same weights as in-app-only users---the model learns an \textit{activity-level gate} rather than channel affinity. This confirms the CSA adapts to source reliability per user rather than memorizing channel-specific shortcuts. A preliminary offline analysis tests whether activity-level weights would suffice. Replacing each user's CSA weights with the mean weights of their L28 activity bucket and re-scoring without retraining left ranking quality on next-day time-spent unchanged ($|\Delta\text{NDCG}| \leq 0.0001$ over 1.1M users on a single snapshot), at granularities from six buckets to 458 activity $\times$ availability cells. This is suggestive only: ranking quality is not the deployed objective, since the policy consumes only the top-$N$ slots surviving the avoidance window, and whether the simpler parameterization holds up on execution efficiency remains an online question.

\subsubsection{Best Time Coordination Experiment}
To test whether cannibalization across use cases exists in the execution time optimization, we selected 10 push notification use cases (each scheduled once daily), which use the same prediction metric, best time signal assembler, and best time policy in the scheduling requests. In the control condition without coordination, all 10 use cases were scheduled and issued the notification sending requests in the same time slot to each user. In the test condition, we classified 5 use cases as high priority group, and the others as low priority group where we avoided selecting the top peak time slot in the best time policy implementation. We observed that introducing the coordination system not only improves the notifications delivery efficiency in both groups (high priority +0.13\%, low priority +0.7\%) but also increases the global product impact (+0.022\%) in Table ~\ref{besttimecoordination}.

\begin{table}[t]
	\centering
	\begin{tabular}{l|cc} \toprule & High priority & Low priority \\ \hline Efficiency (\%) & +0.13$\pm$0.02 & +0.70$\pm$0.12 \\ Impact (\%) & \multicolumn{2}{c}{+0.022$\pm$0.016} \\ \bottomrule \end{tabular}
	\caption{Delivery efficiency and global product impact of the tiered coordination system, relative to a no-coordination baseline. The larger low-priority gain is expected: without coordination all ten use cases collided on the same peak slot, where lower-priority notifications were most crowded out. Shifting them off-peak removes the largest source of cannibalization loss, whereas the high priority group benefits only from the reduced competition. Impact is the relative app-level daily active user change. Values after $\pm$ are 95\% CI half-widths and all results have $p < 0.05$.}
	\label{besttimecoordination}
\end{table}

\subsection{Deployment Lessons} 
\subsubsection{\textbf{A variable-length day broke an implicit interface contract.}} 

Slot indices are defined in a single server-side timezone, because daylight saving transition dates differ across countries, making per-user local-hour bucketing inconsistent globally; the transition itself is handled at the modeling level through a ranking feature flagging countries in transition. Nevertheless, on a transition day the server-side day spans 23 or 25 hours while the offline inference pipeline emits a fixed 24-element array, and the resulting size mismatch raised an unhandled exception in the online service, which we resolved by normalizing every day to exactly 24 slots. The lesson generalizes beyond timezones: when an offline pipeline and an online service each encode a calendar assumption independently, the assumption must be an explicit, validated contract rather than a shared convention, especially for conditions that recur only twice a year and therefore escape routine testing.

\subsubsection{\textbf{Timing headroom is channel-dependent.}} Email notifications showed no measurable gain over the heuristic baseline (Table~\ref{result-table}), which we attribute to
users batch-processing email at self-chosen times rather than responding to individual arrival times. The learned utility adds most where arrival time competes with a narrow window of attention; where it does not, a peak-hour rule is already close to whatever headroom exists at matched scheduling volume.

\section{Conclusion and Future Work}
\label{others}
We presented a deployed execution time optimization system for scheduled jobs, in production for over four years and serving billions of users daily across hundreds of product use cases. It combines learning-to-rank with a best time policy to select personalized execution timestamps, in a compositional architecture where shared user-level activity signals are building blocks for use-case-specific optimization. We traced its evolution from pointwise activity pattern signals, to a Linear Signal Assembler with static global weights, to a Contextual Signal Assembler producing per-user adaptive fusion weights via listwise learning-to-rank.

Through four years of deployment, we accumulated several generalizable insights: 
(1) timing headroom is channel-dependent: against a heuristic peak-hour baseline we measured gains on push, red alert, and SMS notifications, but no measurable gain on email in the use case we tested, where the heuristic was already close enough that we could not resolve a difference; (2) the best timing signal does not always come from the delivery channel itself: in the SMS use case we tested, the in-app time-spent pattern outperformed notification CTR as the training label; (3) cross-use-case cannibalization emerges when multiple schedulers independently optimize for the same peak hours, which we measured across ten push use cases sharing one channel; and (4) a compositional architecture that separates shared signal production from use-case-specific optimization enables incremental adoption at low marginal computational cost. To our knowledge, this represents the first ML-based multi-tenant execution time optimization system deployed across different product domains at industrial scale. 

Three limitations define our next steps. First, the best time policy remains rule-based: $w$ is configured per use case rather than learned, so every user within a use case receives the same spacing. Since selections are interdependent and no optimal slot set is known to supervise against, we plan to personalize the policy instead, predicting $w$ from each user's temporal activity pattern. Second, the coordination layer still resolves contention with hand-assigned priority tiers and no learning loop; we are pursuing a learned tier allocation that optimizes directly for platform-level value and adapts as the tenant mix changes. Finally, the contextual assembler's online validation covers a single use case at one execution per day, and the offline evidence that its weights reduce to activity-level buckets still awaits online confirmation; both are open.

\section{Acknowledgments}
We thank Yuankai Ge for his support of this project. We thank all partners and client teams' collaboration and contributions.

\bibliography{ref}
\end{document}